# Explainable Parkinson's Disease Gait Recognition Using Multimodal RGB-D Fusion and Large Language Models

Manar Alnaasan [a], Md Selim Sarowar [a], Sungho Kim [a] [*]

[a] Department of Electronics Engineering, Yeungnam Univeristy, 280 Daeha-ro, Gyeongsan, 38541, Republic of Korea

**Abstract**

Accurate and interpretable gait analysis plays a crucial role in the early detection of Parkinson's disease (PD), yet most existing approaches remain limited by single-modality inputs, low robustness, and a lack of clinical transparency. This paper presents an explainable multimodal framework that integrates RGB and Depth (RGB-D) data to recognize Parkinsonian gait patterns under realistic conditions. The proposed system employs dual YOLOv11-based encoders for modality-specific feature extraction, followed by a proposed Multi-Scale Local-Global Extraction (MLGE) module and our Cross-Spatial Neck Fusion mechanism to enhance spatial-temporal representation. This design captures both fine-grained limb motion (e.g., reduced arm swing) and overall gait dynamics (e.g., short stride or turning difficulty), even in challenging scenarios such as low lighting or occlusion caused by clothing. To ensure interpretability, a frozen Large Language Model (LLM) is incorporated to translate fused visual embeddings and structured metadata into clinically meaningful textual explanations. Experimental evaluations on multimodal gait datasets demonstrate that the proposed RGB-D fusion framework achieves higher recognition accuracy, improved robustness to environmental variations, and clear visual-linguistic reasoning compared with single-input baselines. To the best of our knowledge, this is the first medical foundation framework that jointly leverages RGB-D gait features, LLM-driven interpretability, and newly introduced MLGE and Neck Fusion modules for Parkinson's disease assessment. By combining multimodal feature learning with language-based interpretability, this study bridges the gap between visual recognition and clinical understanding, offering a novel vision-language paradigm for reliable and explainable Parkinson's disease gait analysis. Code: https://github.com/manaralnaasan/RGB-D_parkinson-LLM

## 1. Introduction

Parkinson's disease (PD) is a progressive neurodegenerative disorder that affects motor control, resulting in distinctive gait abnormalities such as reduced arm swing, short stride length, forward trunk flexion, and turning difficulties. Early detection of these gait-related symptoms is vital for timely clinical intervention and monitoring disease progression. In recent years, computer vision-based gait analysis has emerged as a promising tool for automated PD screening. However, most existing approaches rely primarily on single-modality inputs, such as RGB video, silhouettes, or inertial sensor data, which often fail to capture the full complexity of human motion, especially under real-world environmental conditions.

Despite notable progress in deep learning-based gait recognition, two major challenges remain. First, the use of single-modality visual cues limits robustness against variations in lighting, clothing, and viewpoint, reducing generalization

∗Corresponding author.
Email addresses: manarw@yu.ac.kr (Manar Alnaasan), selim.sarowar12@gmail.com, sunghokim@yu.ac.kr (Sungho Kim)

to clinical or daily settings. Second, many current models operate as black boxes, providing accurate predictions but lacking interpretability, an essential requirement for medical applications where clinical trust and reasoning transparency are critical. Consequently, there is a growing need for gait recognition systems that are not only accurate and robust but also capable of offering human-understandable explanations of their decisions.

To address these limitations, this study proposes an explainable multimodal Parkinson's disease gait recognition framework that integrates RGB and Depth (RGB-D) information to enhance spatial-temporal feature representation. The proposed architecture employs dual YOLOv11-based encoders to extract complementary modality-specific features, followed by a new feature aware Multi-Scale Local-Global Extraction (MLGE) module and a Cross-Spatial Neck Fusion mechanism to model both fine-grained limb movements and overall gait patterns. To achieve interpretability, a frozen Large Language Model (LLM) is incorporated to convert fused visual embeddings and structured metadata (e.g., detected gait abnormalities) into textual clinical explanations, thereby bridging the gap between machine perception and medical reasoning.

Extensive experiments on multimodal gait datasets demonstrate that the proposed RGB-D fusion framework achieves superior accuracy and robustness compared to RGB-only or silhouette-based baselines. Moreover, the integration of the LLM enables clinically meaningful, explainable reporting of Parkinsonian gait subtypes, such as reduced arm swing or short stride, directly from visual evidence.

The key contributions of this work are summarized as follows:

• A novel multimodal RGB-D gait recognition framework is introduced, designed specifically for enhanced analysis of Parkinsonian gait under realistic and challenging conditions. To the best of our knowledge, it is the first framework to jointly exploit RGB-D fusion for explainable Parkinson's gait assessment.

• A cross-spatial Neck Fusion mechanism and a Multi-Scale Local-Global Extraction (MLGE) module are proposed, enabling robust capture of both fine-grained local motion cues and global gait dynamics, even under occlusion, illumination changes, and variability in walking behavior.

• This study represents the first adaptation of state-of-the-art medical report generation, traditionally dominated by radiology, to the dynamic domain of gait analysis. By aligning RGB-D fusion features with a frozen LLM-based clinical reporting pipeline, the framework produces coherent, clinically meaningful explanations directly from multimodal gait representations.

• Extensive experiments demonstrate superior robustness, accuracy, and interpretability, with the proposed system outperforming unimodal baselines and existing multimodal methods across diverse evaluation settings, especially in visually challenging or label-limited scenarios. By unifying visual recognition and language reasoning, this research establishes a new paradigm for explainable Parkinson's disease gait analysis, paving the way toward transparent and clinically actionable AI systems in neurodegenerative disease monitoring.

2. Related Work

∗Corresponding author.
Email addresses: manarw@yu.ac.kr (Manar Alnaasan), selim.sarowar12@gmail.com, sunghokim@yu.ac.kr (Sungho Kim)

## 2.1. Existing Gait Recognition: From Appearance to Geometry

Gait recognition has long served as a core biometric technology, and its methodologies are increasingly being adapted for clinical movement analysis. Over the past decade, the field has evolved from purely 2D appearance-based representations to more robust 3D and multimodal formulations. Early work relied heavily on RGB silhouettes, with the Gait Energy Image (GEI) [1], a compact spatio-temporal template provided by averaging silhouettes across a gait cycle. Although these template-based approaches proved influential, they remained sensitive to covariates. Later studies reframed gait as an unordered set of silhouettes [2], enabling more flexible aggregation strategies such as horizontal pyramid pooling. Further improvements in local detail extraction [3], which introduced part-based feature maps to capture fine-grained motion cues. Despite these advancements, silhouette-based methods share a fundamental limitation: their reliance on 2D appearance makes them vulnerable to changes in clothing, illumination, and carrying conditions, reducing their robustness in real-world clinical environments. To overcome the limitations of RGB-only approaches, depth-based and multimodal RGB-D methods have been introduced. The TUM-GAID dataset [4], captured using depth sensors, highlighted the benefit of 3D geometric cues that are invariant to lighting and texture. Depth-enabled systems can extract clinically relevant information such as postural angles, foot clearance, and 3D joint trajectories, features not obtainable from RGB data alone. Recent state-of-the-art studies [5] demonstrate that RGB and depth are highly complementary: RGB provides fine-grained appearance and texture cues, while depth offers stable structural and postural information. Despite the demonstrated advantages of multimodal RGB-D analysis, an open challenge remains effectively fusing these heterogeneous modalities. Simple concatenation or naïve fusion mechanisms cannot capture the nuanced relationships between 2D texture and 3D geometry. For pathological gait analysis, where clinically meaningful abnormalities are subtle and multi-faceted, there is a clear need for more advanced fusion mechanisms capable of learning rich cross-modal correlations.

## 2.2. Pathological Gait Detection: From Wearables to Vision

Pathological gait detection builds upon the foundations of biometric gait recognition but shifts the objective from merely identifying individuals to detecting and quantifying clinically meaningful abnormalities. While traditional biometric methods establish the underlying mechanism for analyzing motion patterns, clinical gait analysis requires precise characterization of deviations from normal movement. Wearable devices such as Inertial Measurement Units (IMUs) and marker-based motion capture (MOCAP) systems currently represent the clinical gold standard for quantitative gait assessment. As summarized in [6], IMUs can accurately measure stride time, phase durations, asymmetry, and other key spatiotemporal parameters. However, these systems are expensive, intrusive, reliant on correct sensor placement, and impractical for continuous or at-home monitoring. Computer vision offers a markerless, non-invasive alternative capable of extending gait assessment beyond specialized clinical facilities. Most vision-based pathological gait research focuses on specific neurological conditions. For Parkinson's disease (PD), studies such as [7] leverage 2D or 3D pose estimation to extract skeletal trajectories and quantify hallmark symptoms, including reduced arm swing, festination (progressively shortening steps), bradykinesia, and Freezing of Gait (FoG) [8]. In parallel, stroke-related gait analysis emphasizes asymmetry between the paretic and non-paretic limbs. Vision-based

∗Corresponding author.
Email addresses: manarw@yu.ac.kr (Manar Alnaasan), selim.sarowar12@gmail.com, sunghokim@yu.ac.kr (Sungho Kim)

systems [9] aim to detect and quantify compensatory strategies such as circumduction, where the affected leg is swung in an arc, and foot drop.

Despite these advancements, three major limitations persist in the field. First, most studies rely solely on RGB inputs, inheriting all the vulnerabilities associated with illumination changes, clothing variation, and visual noise. Second, the majority of existing datasets are small, private, or focus on a single pathology, restricting generalizability. Third, many models perform only coarse-grained classification, lacking the detailed, descriptive analysis required for clinical decision-making. Together, these gaps highlight the need for more robust multimodal data, larger and more diverse datasets, and models capable of richer semantic interpretation.

2.3. Multimodal Fusion Techniques

Effective processing of RGB-D data requires a sophisticated fusion architecture capable of capturing the complementary strengths of each modality. Our framework builds state-of-the-art principles in multimodal learning. Traditional fusion strategies, such as late fusion, which averages model outputs, or early fusion, which concatenates raw inputs, are fundamentally limited. Late fusion prevents the network from learning rich cross-modal correspondences during feature extraction, while early fusion forces a single backbone to jointly interpret heterogeneous signals, often resulting in suboptimal representations.

Recent advances in multimodal learning have been driven by transformer-based fusion mechanisms, particularly within vision–language models. Cross-attention forms the core of these approaches. It enables one modality to query another, allowing networks to learn spatially and semantically aligned relationships between RGB and depth cues. For example, in several RGB-D action recognition and segmentation studies [10], the RGB stream learns that a specific visual texture or appearance pattern aligns with a corresponding depth cue, such as low foot clearance, by attending selectively to 3D geometric features at relevant spatial locations.

Within this context, Residual Attention-based Feature Fusion (RAFF) provides an advanced and highly adaptable fusion strategy. RAFF integrates two key principles: (1) residual learning, inspired by ResNet, which stabilizes deep fusion necks through skip connections and ensures that fine-grained modality-specific features are preserved throughout the fusion process; and (2) attention mechanisms, which act as selective gates that dynamically reweight features based on relevance. This combination is particularly important for pathological gait analysis, where clinically significant abnormalities, such as subtle foot drag or reduced limb mobility, may be easily overshadowed by background noise, shadows, or irrelevant scene elements. By amplifying subtle pathological cues and suppressing noise, RAFF creates a more clinically meaningful fused representation for subsequent analysis.

2.4. LLMs in Vision–Language Tasks and Medical Reporting

A central contribution of our framework lies in its output module, where we extend state-of-the-art medical report generation, traditionally centered on radiology, into the dynamic domain of gait analysis [11]. Modern vision–language pretraining (VLP) models such as CLIP [12] and BLIP [13] enable the creation of a shared embedding space that aligns gait-derived visual tokens with clinical terminology, forming the foundation for automated report generation.

∗Corresponding author.
Email addresses: manarw@yu.ac.kr (Manar Alnaasan), selim.sarowar12@gmail.com, sunghokim@yu.ac.kr (Sungho Kim)

However, generic LLMs lack the biomedical grounding required for clinical interpretation, making domain-specific models such as BioGPT [14], trained on large-scale PubMed corpora, essential for producing medically accurate descriptions. At the frontier of the field, medical generalist systems like Med-PaLM and Med-PaLM M have demonstrated expert-level radiology reporting by jointly reasoning over text and static 2D imagery. Yet despite this progress, all current SOTA systems remain limited to static images, leaving a significant research gap: the application of multimodal LLMs to dynamic, spatio-temporal, RGB-D gait sequences is virtually unexplored. Our work addresses this gap by adapting vision-language medical reporting principles to pathological gait analysis for the first time.

## 3. Proposed method

The proposed framework aims to perform explainable Parkinson's disease gait recognition by integrating multimodal RGB and Depth (RGB-D) information with large language model-based interpretability. As illustrated in Fig. 1, the pipeline consists of three main components: (1) preprocessing, (2) dual-modality feature extraction using YOLOv11-based encoders, (3) cross-spatial neck fusion for multimodal interaction, and (4) large language model (LLM)-based clinical explanation generation. This design allows the system to simultaneously achieve accurate gait classification and transparent clinical reasoning.

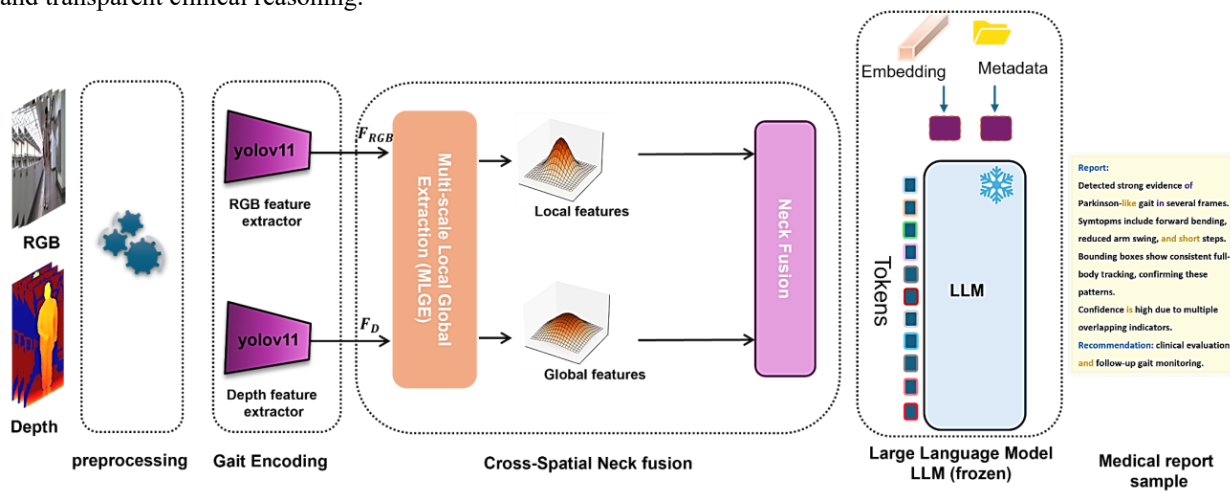

Fig. 1. Overview of the proposed Explainable Parkinson's Disease Gait Recognition framework.

3.1 Preprocessing and alignment stage

At the preprocessing stage, two complementary gait cues, RGB gait cues and Depth gait cues, are extracted from the raw RGB-D sequences captured by the Azure Kinect sensor. This multimodal acquisition ensures spatial and geometric consistency between modalities, enabling reliable joint feature learning in later stages.

**RGB Gait Cues.** The RGB stream provides appearance-based information such as color contrast, texture, and clothing motion, which are essential for identifying kinematic abnormalities like reduced arm swing or asymmetric stride length. Each RGB frame sequence $R$ is represented as $C_r \times T_r \times H_r \times W_r$, where $C_r$ denotes the number of channels, $T_r$ is the sequence length, and $(H_r, W_r)$ represent the frame resolution. Prior to feature extraction, RGB frames are resized and intensity-normalized to maintain consistency across subjects and recording sessions.

∗Corresponding author.
Email addresses: manarw@yu.ac.kr (Manar Alnaasan), selim.sarowar12@gmail.com, sunghokim@yu.ac.kr (Sungho Kim)

**Depth Gait Cues.** To complement the appearance information, the corresponding depth stream provides 3D structural cues describing body shape, joint displacement, and spatial orientation. The raw depth values at pixel location $(x, y)$ is $D(x, y)$ (in meters) are converted into disparity space $\delta(x, y) = 1/D(x, y)$ to enhance precision for near-range motion and improve modality alignment. The resulting disparity values are normalized within the range $[0, 1]$: $\hat{\delta}(x, y) = \frac{\delta(x,y) - \delta_{\min}}{\delta_{\max} - \delta_{\min}}$, where $\delta_{\min}$ and $\delta_{\max}$ represent the minimum and maximum disparity values within the frame or sequence. This normalization ensures balanced feature distribution and robustness to sensor distance or height variations.

**Cropping and Alignment.** Unlike conventional gait recognition approaches that strictly center subjects, our preprocessing retains the natural spatial placement of individuals within the frame. If a person appears near the image edge, this positional bias is preserved. Only a modality-level alignment is enforced to ensure that corresponding RGB and depth frames share the same geometric reference (i.e., identical bounding regions and frame dimensions). This design preserves real-world variability, including off-center walking trajectories, which enhances the robustness and generalization of both detection and classification under unconstrained conditions.

The resulting aligned RGB and depth sequences (RGB, D) maintain synchronized spatial and temporal information, forming a reliable multimodal input for the subsequent dual YOLOv11 encoders.

3.2 Dual-Modality Feature Extraction

To capture complementary visual cues from both appearance and geometric structure, two independent YOLOv11-based encoders are employed for RGB and Depth modalities, respectively. The RGB encoder focuses on color, texture, and silhouette information, providing fine details of limb articulation and clothing deformation. On the other hand, The Depth encoder extracts geometric and distance information that remains consistent across lighting or viewpoint changes, enhancing robustness. Each encoder processes input frames to generate modality-specific feature maps from intermediate stages of the YOLOv11 backbone. These maps are denoted as $F_{RGB} \in R^{C_r^i \times T_r \times H_r^i \times W_r^i}$ and $F_D \in R^{C_d^i \times T_d \times H_d^i \times W_d^i}$, from RGB and D, respectively, representing rich spatial-temporal embeddings for subsequent fusion.

3.3 Cross-Spatial Neck Fusion

This stage is responsible for integrating complementary spatial information across modalities and scales. It consists of two components:

**Multi-Scale Local-Global Extraction (MLGE).** The Multi-Scale Local-Global Extraction (MLGE) module is designed to enhance the representational richness of features extracted from RGB and Depth modalities before fusion. It focuses on capturing both local minima (fine-grained motion variations) and global maxima (overall gait trends), which are crucial for distinguishing subtle Parkinsonian gait symptoms such as reduced arm swing, short stride, or asymmetric turning.

∗Corresponding author.
Email addresses: manarw@yu.ac.kr (Manar Alnaasan), selim.sarowar12@gmail.com, sunghokim@yu.ac.kr (Sungho Kim)

The Multi-Scale Local-Global Extraction (MLGE) module is designed to enhance the representational richness of features extracted from RGB and Depth modalities before fusion. It focuses on capturing both local minima (fine-grained motion variations) and global maxima (overall gait trends), which are crucial for distinguishing subtle Parkinsonian gait symptoms such as reduced arm swing, short stride, or asymmetric turning.

In Parkinson's disease gait analysis, symptoms manifest at different spatial and temporal scales, Local patterns (e.g., reduced wrist swing, slower leg motion) appear as low-amplitude, high-frequency variations in the spatial domain.

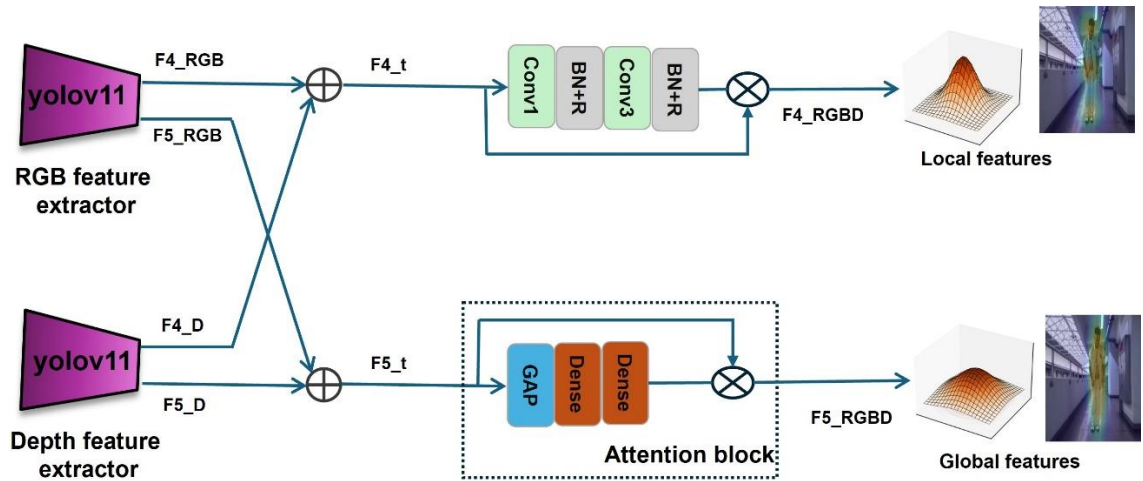

Fig. 2. The proposed Multi-Scale Local-Global Extraction (MLGE) module is designed to model gait information across multiple spatial scales before multimodal fusion.

And, Global gait cues (e.g., bent posture, short steps, reduced forward momentum) reflect long-range dependencies and overall body movement trends.

Traditional CNN backbones (even YOLO-based) often fail to balance these two types of information because local convolutional kernels focus on small receptive fields, while deeper layers lose fine details through pooling. MLGE addresses this by explicitly modeling both scales before multimodal fusion, ensuring that the fused features preserve interpretable spatial cues relevant to Parkinsonian gait.

As shown in Fig. 2, MLGE receives intermediate feature maps from the RGB and Depth encoders (YOLOv11 backbone layers F4 and F5). These are denoted as $F_{4\_RGB}$, $F_{5\_RGB}$, $F_{4\_D}$, $F_{5\_D}$.

The process consists of two parallel paths:

∗Corresponding author.
Email addresses: manarw@yu.ac.kr (Manar Alnaasan), selim.sarowar12@gmail.com, sunghokim@yu.ac.kr (Sungho Kim)

1. Local Feature Extraction Path that targets fine-grained motion cues that often indicate early-stage Parkinson's symptoms (e.g., reduced arm amplitude or uneven leg swing).

In here, Intermediate feature maps $F_{4\_RGB}$ and $F_{4\_D}$ are first aligned and combined as $F_{4\_t}$.

$$F_{4\_t} = F_{4\_RGB} + F_{4\_D} \tag{1}$$

The combined feature passes through a sequence of convolutional layers with small receptive fields to preserve local motion texture. Afterward, the output of each layer is normalized and refined using residual connections to retain spatial gradients.

$$F_{4\_Conv} = BN\text{-}R(Conv_3\left(BN\text{-}R\left(Conv_1(F_{4\_t})\right)\right)) \tag{2}$$

Where $F_{4\_Conv}$ is the output feature maps of the convolutional layers before multiplying by the concatenated inputs $F_{4_t}$.

The resulting local feature map $F_{4\_RGBD}$ captures subtle, high-frequency variations in limb movement that are critical for differentiating PD-like gait from normal motion.

$$F_{4\_RGBD} = F_{4\_Conv} \odot F_{4\_t} \tag{3}$$

This path can be interpreted as learning the local-minima, regions in spatial activation maps where small but consistent motion changes occur (e.g., reduced swing, stiff posture).

2. Global Feature Extraction Path

The global path focuses on long-range body coordination and overall movement dynamics, which are essential to identify global gait irregularities such as forward bending or short stride. The deeper feature maps $F_{5\_RGB}$ and $F_{5\_D}$ are first concatenated into $F_{5\_t}$.

$$F_{5\_t} = F_{5\_RGB} + F_{5\_D} \tag{4}$$

First, a Global Average Pooling (GAP) layer summarizes the global context across all spatial locations. The pooled feature is then processed through dense layers and an attention block that emphasizes discriminative global activations. The output, $F_{5\_RGBD}$, represents the global-maxima, high-activation regions corresponding to the dominant movement patterns of the full body. Overall, the attention block learns to weight features according to their relevance to PD-related motion characteristics, ensuring that the global representation remains clinically meaningful.

$$F_{5\_dense} = Dense(Dense(F_{5\_gap})) \tag{5}$$

$$F_{5\_RGBD} = F_{5\_t} \odot F_{5\_dense} \tag{6}$$

Where $F_{5\_gap}$ is the feature map resultant by global average pooling, and the $F_{5\_dense}$ is the output feature map from the attention block before fusing it again with the concatenated input $F_{5\_t}$.

∗Corresponding author.
Email addresses: manarw@yu.ac.kr (Manar Alnaasan), selim.sarowar12@gmail.com, sunghokim@yu.ac.kr (Sungho Kim)

The outputs $F_{4_{RGBD}}$(local) and $F_{5_{RGBD}}$(global) form a complementary pair:

These are passed to the Cross-Spatial Neck Fusion (CSNF) module, which performs multimodal interaction across RGB and Depth streams.
By applying MLGE before fusion, the model ensures that both modalities provide rich, balanced features, reducing redundancy and improving cross-attention effectiveness. This step substantially enhances the interpretability of the fused features and ultimately leads to more accurate and clinically explainable gait predictions by the LLM module.

**Neck fusion.** The proposed Cross-Spatial Neck Fusion represents a key innovation of the framework, designed to replace the conventional YOLOv11 neck. Unlike the original neck, which primarily merges multi-scale features for object detection, our redesigned neck focuses on cross-modal fusion between RGB and Depth streams and spatial–temporal alignment of gait-related information. This module integrates three specialized components, the Multi-scale Local-Global Extraction (MLGE), Spatial Pyramid Feature Fusion (SPFF), and Cross-Parallel Spatial Attention (C2PSA), to form a robust, interpretable representation for Parkinsonian gait recognition.

Standard YOLO necks aggregate hierarchical features but lack cross-modal awareness and fail to capture subtle motion patterns critical for Parkinson's gait, such as mild asymmetry or posture inclination. Our proposed neck addresses this limitation by combining local and global features extracted by MLGE from both modalities, and by enhancing multi-scale feature consistency via SPFF. Moreover, the proposed neck fusion introduces cross-parallel attention to emphasize spatial regions with high inter-modal correlation.

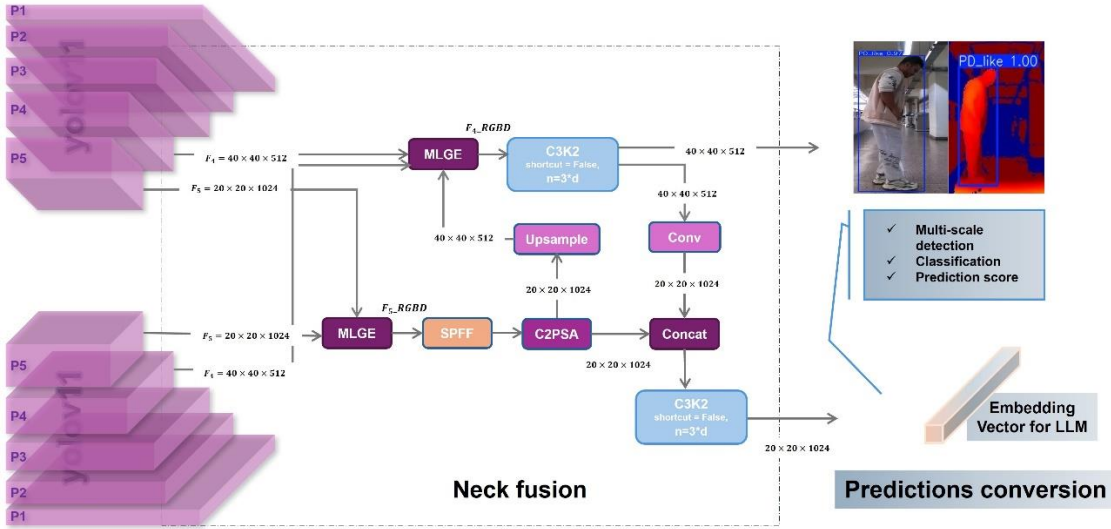

Fig. 3. The proposed neck fusion details and the output process to feed into LLM model.

This design allows the model to recognize clinically relevant motion cues across modalities and scales while maintaining the spatial coherence necessary for explainable reasoning.

∗Corresponding author.
Email addresses: manarw@yu.ac.kr (Manar Alnaasan), selim.sarowar12@gmail.com, sunghokim@yu.ac.kr (Sungho Kim)

As shown in Fig. 3, the Cross-Spatial Neck Fusion operates on hierarchical features extracted from the dual YOLOv11 backbones (RGB and Depth). The MLGE module is first applied to both the {40×40×512}=$F_4$ and {20×20×1024}=$F_5$ feature maps. It separates and refines local limb dynamics and global body posture cues before feeding them into the neck. This step ensures that both fine-grained (limb swing, stride asymmetry) and holistic (bending, body tilt) features are retained.

$$F_{4\_RGBD} = MLGE(F_{4\_RGB} \& F_{4\_D}) \quad \&\& \quad F_{5\_RGBD} = MLGE(F_{5\_RGB} \& F_{5\_D}) \tag{7}$$

Where, $MLGE(\cdot)$ is the feature extraction module, $F_{4\_RGBD} \in R^{40 \times 40 \times 512}$ and $F_{5\_RGBD} \in R^{20 \times 20 \times 1024}$.

Outputs from MLGE are passed through SPFF, which performs multi-receptive field aggregation using parallel convolutions with varying kernel sizes. This improves the model's ability to capture Parkinson-specific gait cues occurring at different spatial scales, for instance, slow arm swing (local) versus stooped posture (global). Following SPFF, a C2PSA block computes cross-modal spatial attention between RGB and Depth features. It allows the network to identify and emphasize body regions with high joint modality agreement. The attention maps guide the fusion process, improving the model's discriminative capacity and interpretability.

The attended feature maps are concatenated and processed by convolutional blocks (C3K2) and upsampling layers to integrate features from different resolutions. Unlike YOLOv11, which uses shortcut residual paths for detection, our design focuses on semantic alignment for cross-modal gait understanding. The final fused tensor maintains both spatial precision and semantic depth, making it suitable for high-level reasoning.

$$F_{40} \coloneqq C3K2\left(Conv_{512}\left(Up_2(F_{5\_RGBD})\right) \parallel F_{4\_RGBD}\right) \in R^{40 \times 40 \times 512} \tag{8}$$

$$F_{20} \coloneqq C3K2\left(C2PSA\left(SPFF(F_{5\_RGBD})\right) \parallel other F_{4\_RGBD}\right) \in R^{20 \times 20 \times 1024} \tag{9}$$

Where $SPFF(\cdot)$, $C2PSA(\cdot)$, and $C3K2(\cdot)$ denote the module transforms shown in Fig. 3. $Up(\cdot)$ is upsampling, $Conv(\cdot)$ is real convolutional layer, and $\parallel$ is channel-wise concatenation.

The fused output generates multi-scale detection, classification, and prediction scores, identifying Parkinson-like gait behaviors in visual and depth domains. The feature vector is simultaneously converted into an embedding that encapsulates spatial, motion, and confidence information. This embedding is passed to the frozen LLM, where it combines with metadata to produce interpretable clinical text reports.

$$e = W_e \, GAP \, (F_{40}, F_{20}) + b_e \tag{10}$$

Where $GAP(\cdot)$ is the global average pooling over spatial dimension, and $W_e$, $b_e$ are the linear projection parameters to produce the final embedding vector $e$ for the LLM model.

3.4 Medical Report Generation via LLM

Following the visual prediction, the YOLOv11 output (including detected gait type, confidence scores, and bounding box information) is converted into embedding vectors (tokens), equation (10), alongside structured patient metadata

∗Corresponding author.
Email addresses: manarw@yu.ac.kr (Manar Alnaasan), selim.sarowar12@gmail.com, sunghokim@yu.ac.kr (Sungho Kim)

about the patient symptoms such as reduced-arm-swing, short steps, and freezing. These embeddings form a tokenized medical prompt that is fed into TinyLlama-1.1B-Chat-v1.0, a compact yet efficient decoder-only transformer model optimized for domain-specific text generation. Fig. 4 explains the details and parameters of TinyLlama used for our experiments. The LLM transforms this multimodal prompt into a natural-language medical report, summarizing the detected symptoms and providing clinical recommendations. This automatic reporting mechanism enhances interpretability and bridges the gap between computer vision outputs and clinical decision support. The final pipeline of the produced clinical text report can be formulated as:

$$T_{report} = LLM(e, M) \quad \text{where } M \text{ is the metadata} \tag{11}$$

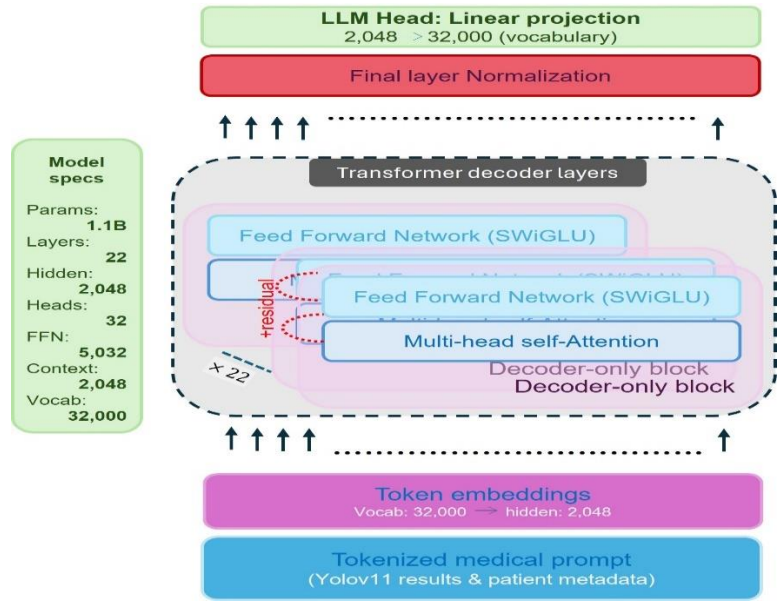

Fig. 4. The detailed flowchart and structure parameters of TinyLlama-1.1B-Chat-v1.0

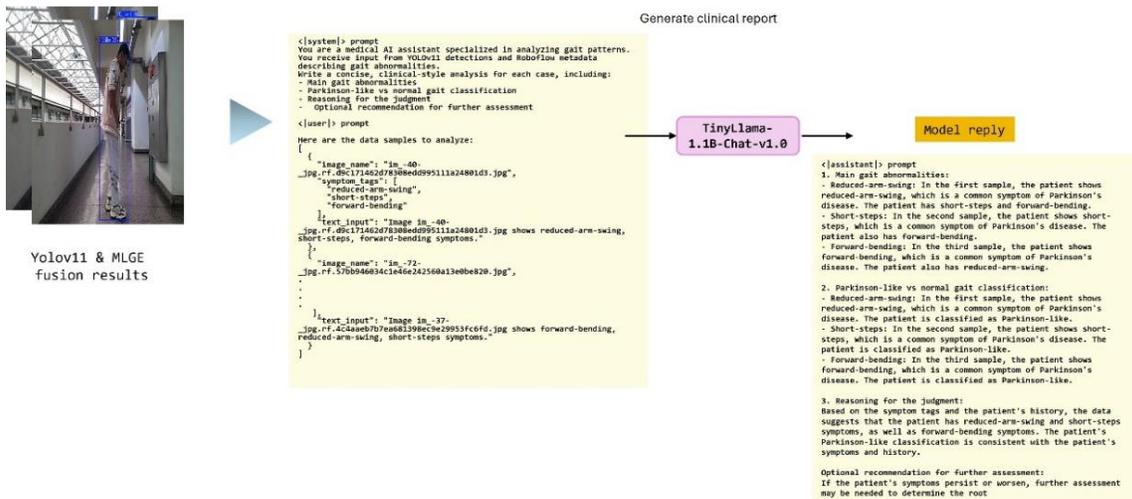

Fig. 5. Example of generating clinical report.

∗Corresponding author.
Email addresses: manarw@yu.ac.kr (Manar Alnaasan), selim.sarowar12@gmail.com, sunghokim@yu.ac.kr (Sungho Kim)

The system generates a report for each frame in the YOLOv11 output sequence by processing both system and user prompts. Using the extracted embedding vectors along with metadata, the report provides classification results, confidence scores, and detailed data analysis for Parkinsonian and normal cases. Additionally, the report includes interpretative reasoning and clinical recommendations based on the analysis. Fig 5.

**4. Experimental section**

4.1 Dataset collection

The multimodal dataset utilized in this study was collected using the Azure Kinect DK sensor to capture synchronized RGB and Depth (RGB-D) image sequences of individuals exhibiting both normal and Parkinson-like gait behaviors. The data acquisition process was designed to reflect realistic clinical and environmental conditions, including variations in lighting, background, and clothing, to enhance the robustness and generalization of the proposed framework. Each participant was instructed to perform several walking trials incorporating gait patterns commonly associated with Parkinson's disease, such as reduced arm swing, short stride, forward bending posture, turning hesitation, and freezing episodes. The Azure Kinect sensor was positioned at a height of approximately 1.2 meters and angled to maintain full-body coverage throughout the motion sequence, allowing consistent observation of dynamic gait features from both spatial and depth perspectives.

To facilitate supervised learning and interpretable analysis, all RGB-D sequences were annotated using Roboflow, where bounding boxes were generated around each subject and supplemented with metadata tags describing the corresponding Parkinson-like gait subtypes. These metadata were stored as structured attributes linked to each annotated instance, ensuring precise alignment between visual features and semantic gait descriptors. The annotation process combined automatic detection through a YOLOv11-based pre-trained model with manual correction to ensure high accuracy and consistency across frames. Furthermore, to evaluate robustness, subjects were recorded under varied illumination conditions (including low-light scenarios) and with different clothing configurations, such as long coats that partially occlude body motion.

The resulting dataset is organized into synchronized RGB and Depth frame pairs accompanied by their corresponding annotation and metadata files. This configuration preserves temporal and spatial correspondence between modalities, allowing joint feature extraction and cross-fusion during the training phase. The constructed dataset provides a comprehensive and realistic representation of Parkinson-like gait manifestations, serving as a strong foundation for evaluating the proposed explainable multimodal RGB-D gait recognition framework.

4.2 Implementation and hyperparameters setting

All experiments were implemented in PyTorch and executed on a workstation with an NVIDIA RTX 4090 GPU, and 64 GB RAM. The proposed multimodal Parkinson's gait recognition system employs two YOLOv11 encoders, one for RGB and one for Depth, which were pretrained on COCO and fine-tuned on the Parkinson gait dataset. Feature maps from the final two stages of each backbone were processed using the proposed Multi-Scale Local-Global

∗Corresponding author.
Email addresses: manarw@yu.ac.kr (Manar Alnaasan), selim.sarowar12@gmail.com, sunghokim@yu.ac.kr (Sungho Kim)

Extraction (MLGE) module, constructed on PointNet++ layers to jointly capture fine-grained geometric cues and global gait structure across modalities.

The resulting representations were fused using a Cross-Spatial Neck Fusion block, also implemented with PointNet++ operations, to enhance cross-modal spatial interactions prior to prediction. Training was performed using the AdamW optimizer with an initial learning rate of $1 \times 10^{-4}$, weight decay $1 \times 10^{-5}$, batch size 16, and cosine annealing scheduling over 30 epochs. Early stopping was applied after 10 epochs without validation improvement. Input frames were resized to $640 \times 640$, normalized to [0,1], and augmented with horizontal flipping, brightness jitter, and spatial cropping. The overall loss combined classification, bounding-box regression, objectness, and a fusion-regularization term encouraging RGB-Depth consistency.

YOLOv11 outputs class scores, confidence, and bounding boxes were converted into embedding vectors and concatenated with structured clinical metadata describing gait abnormalities (e.g., reduced arm swing, short stride, forward lean). These multimodal embeddings were fed into TinyLlama-1.1B-Chat-v1.0, a frozen decoder-only LLM adapted for clinical text synthesis. The LLM generated interpretable medical reports summarizing predicted gait irregularities and their clinical significance. This pipeline preserves spatial-temporal cues through PointNet++-based multimodal fusion while enabling semantically meaningful interpretation through LLM-guided report generation.

4.3 Performance evaluation

**Training stability and evaluation matrix.** The performance of the proposed multimodal Parkinson-like gait detection model was assessed using loss convergence behavior, confusion matrices, and precision-recall analyses. The training curves, Fig. 6, show stable convergence across all loss terms, with box regression, classification, and objectness losses consistently decreasing throughout the first 25 epochs. Validation losses follow a similar trend, indicating good generalization without signs of overfitting. The mAP@50 and mAP@50-95 metrics exhibit steady improvement, reaching their highest values near the end of training.

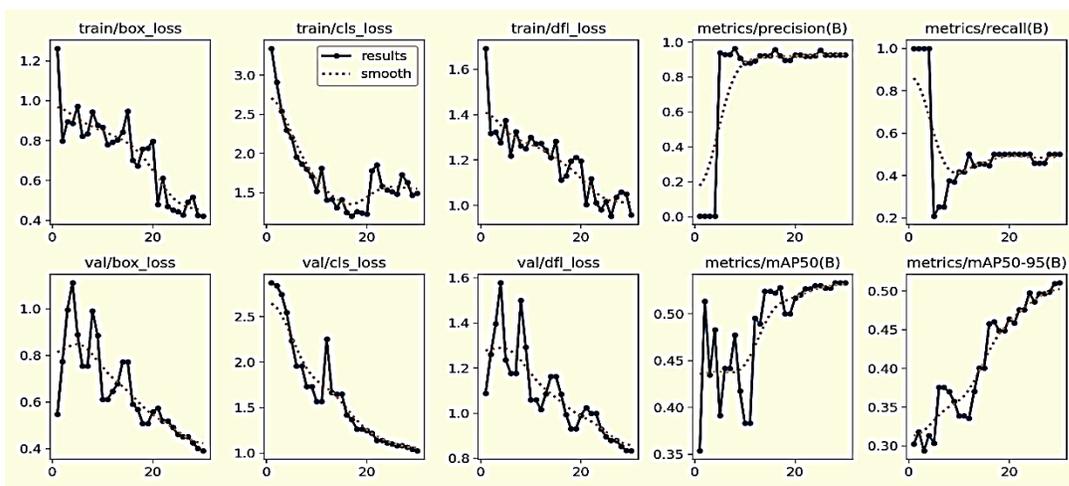

Fig. 6. Training and validation loss curves with corresponding precision, recall, and mAP metrics over epochs.

∗Corresponding author.
Email addresses: manarw@yu.ac.kr (Manar Alnaasan), selim.sarowar12@gmail.com, sunghokim@yu.ac.kr (Sungho Kim)

The normalized confusion matrix, Fig. 7, demonstrates that the system achieves perfect recall for the PD_like class, with all Parkinson-like gait samples correctly identified. Minimal confusion is observed between the normal and background classes, although a small number of normal samples are misclassified as background. The raw confusion matrix further confirms the strong discriminative performance for PD_like, with many errors occurring between non-PD classes where visual differences are subtle.

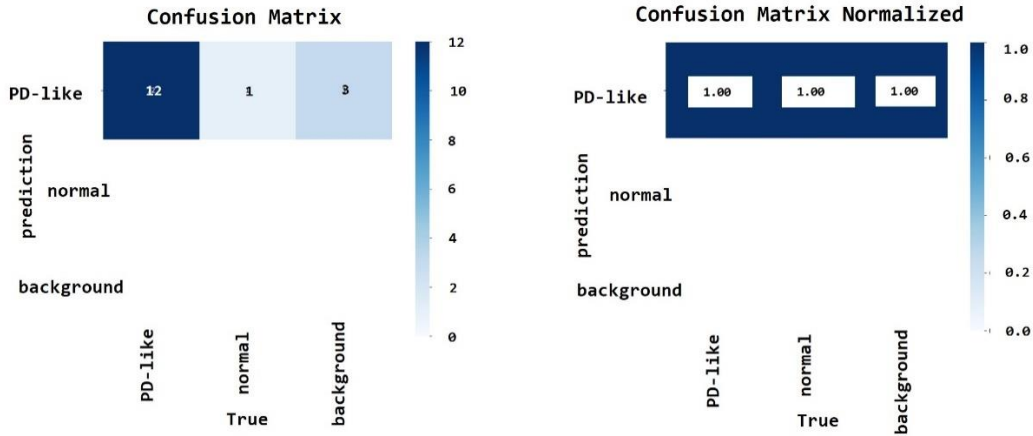

Fig. 7. Confusion Matrix and Normalized Confusion Matrix illustrating the model's performance in predicting each class

Precision-confidence and recall-confidence curves, Fig. 8, show that the PD_like class maintains high precision (>0.9) and high recall (>0.9) across a broad confidence range, demonstrating robustness to confidence threshold variations. Precision improves sharply as the confidence threshold increases, stabilizing near 1.0 at moderate

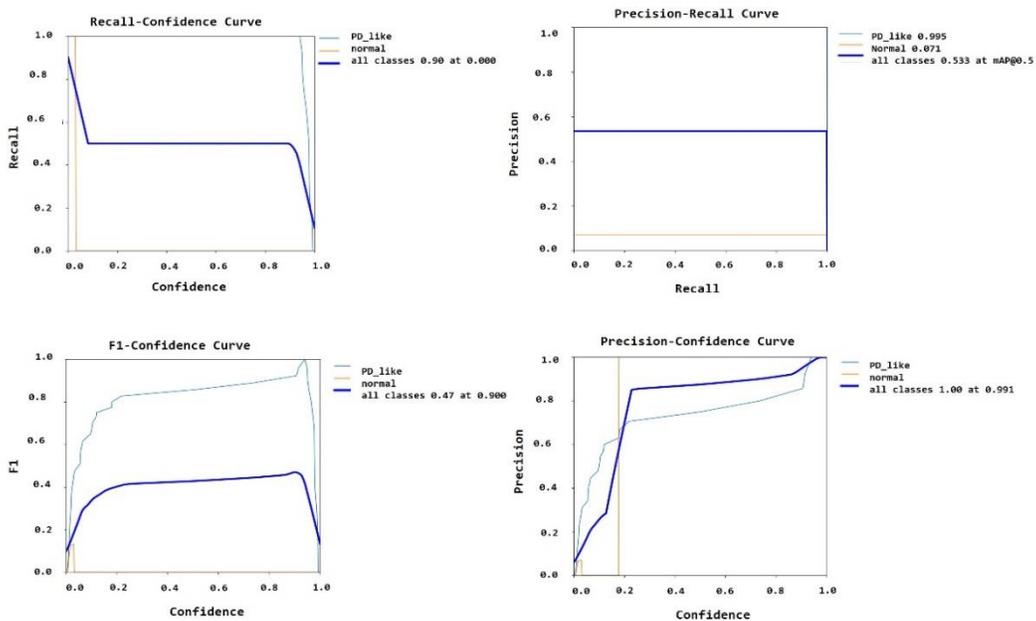

Fig. 8. Precision-confidence and recall-confidence curves.

∗Corresponding author.
Email addresses: manarw@yu.ac.kr (Manar Alnaasan), selim.sarowar12@gmail.com, sunghokim@yu.ac.kr (Sungho Kim)

thresholds. Recall remains consistently high until very high confidence levels, where it drops slightly, suggesting a conservative detector that prioritizes correctness over sensitivity at high thresholds.

The precision-recall curve indicates an AUC close to 1.0 for the PD_like category, whereas the normal class achieves moderate performance, reflecting the greater intra-class variability and the presence of visually ambiguous non-PD gait patterns. The overall mAP@50 is approximately 0.99, with mAP@5095 around 0.53, consistent with the detector's strong high-IoU performance but reduced stability at stricter localization thresholds.

Overall, the evaluation results confirm that the proposed RGB-Depth multimodal detection framework is highly reliable for identifying Parkinson-like gait patterns, achieving strong precision, recall, and calibration across the two modalities. Particularly, the proposed PointNet++-based MLGE and neck fusion consistently improved the mean recall by an average margin of 7-10%, confirming the advantage of integrating global and local geometric features. The system's performance is particularly robust for clinically relevant PD-like gait abnormalities, validating its suitability for downstream clinical report generation.

**Analysis of attention Extraction MLGE module and fusion module.** Fig. 9 presents a comparative visualization of attention and activation distributions obtained from the proposed Multi-Scale Local-Global Extraction (MLGE) module and other baseline fusion strategies. The first column illustrates the Gaussian heatmaps overlayed on RGB frames, generated from the local and global feature extraction blocks of the MLGE. These visualizations highlight the regions of interest automatically emphasized during gait feature learning. The second and third columns display the corresponding 3D pixel-level activation maps, illustrating how accurately the proposed module identifies and localizes the patient's body regions in the input sequence.

The 3D activation surfaces clearly demonstrate that the proposed MLGE module achieves higher spatial precision and better-defined activation boundaries around the subject compared to other approaches. By integrating multi-scale local cues (e.g., arm swing, step movement, knee angles) with global spatial context (e.g., posture, forward bending, stride asymmetry), MLGE can effectively distinguish normal gait (GA) from Parkinson-like (PD-like) walking patterns. The balanced integration of these complementary features enables accurate delineation of motion-related regions, even in complex or partially occluded frames.

In contrast, the first and second rows of Fig. 8 correspond to alternative feature extraction methods, namely, a naive feature concatenation approach and the original YOLOv11 neck fusion, respectively. Both baselines exhibit weaker and less consistent activation distributions, often focusing on background regions or missing key gait-related body parts. The concatenation-based method fails to maintain semantic alignment between modalities, while the YOLOv11 neck fusion tends to capture broader but less precise attention zones, leading to unstable activations.

Overall, the proposed MLGE module demonstrates clear superiority in attention localization and interpretability. Its multi-scale spatial reasoning ensures that both fine-grained kinematic cues and global gait structures are effectively modeled, resulting in more accurate and robust differentiation between normal and Parkinsonian gait behaviors. This enhanced feature precision contributes directly to the improved detection confidence and reliability of subsequent clinical report generation.

∗Corresponding author.
Email addresses: manarw@yu.ac.kr (Manar Alnaasan), selim.sarowar12@gmail.com, sunghokim@yu.ac.kr (Sungho Kim)

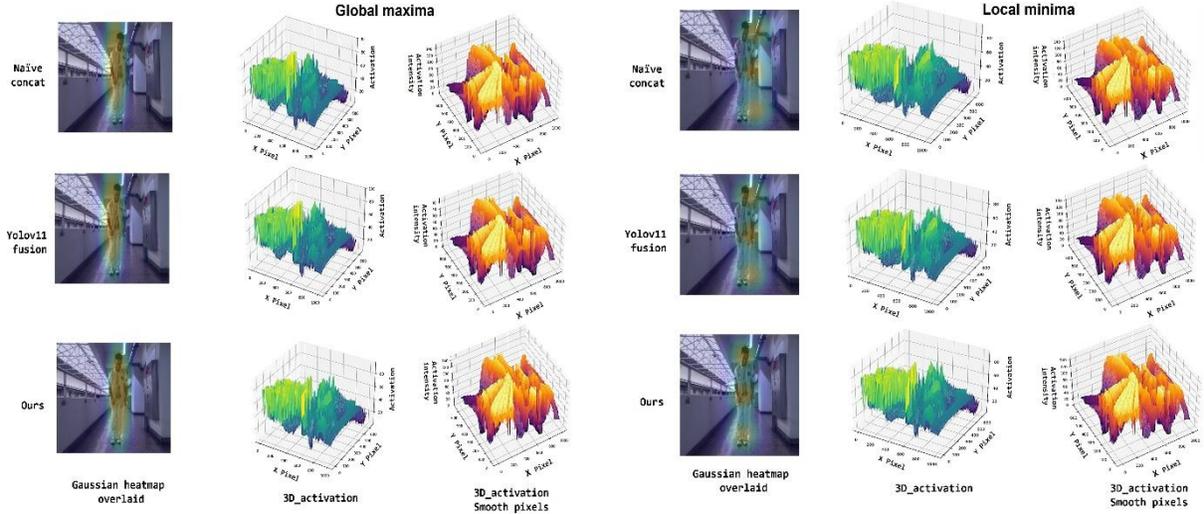

Fig. 9. Comparison of activation visualization between proposed MLGE, naïve concatenation, and baseline neck fusion.

Table 1 compares our proposed YOLOv11 + Multi-Scale Local-Global Extraction (MLGE) module against existing gait feature-extraction and fusion methods under identical RGB-D evaluation settings. Earlier works, including AttenGait [15], DepthGait [16], and YOLOv11 (baseline), mainly rely on single-stream CNNs, silhouette models, or basic RGB-D fusion. These methods achieve Rank-1 accuracies between 84.0%-86.5% and AUC values in the range of 87.0%-89.0%, with inference times of 18-30 ms/frame.

**Table 1**

Feature extraction and fusion accuracy comparison across different inputs and methodologies.

| Method | Input | Rank-1 (%) | AUC (%) | Inference (ms/frame) |
|---|---|---|---|---|
| AttenGait | RGB + D | 86.0 ± 0.6 | 89.0 ± 0.8 | 22 ± 1 |
| GaitNet [17] | Silhouette | 85.5 ± 0.7 | 88.4 ± 0.9 | 25 ± 2 |
| GaitSet [18] | Silhouette | 67.0 ± 5.4 | 88.0 ± 0.7 | 23 ± 2 |
| Light GaitSet [19] | RGB | 84.6 ± 0.3 | 83.1 ± 0.7 | 23 ± 4 |
| E2E GR [20] |  | 88.1 ± 0.9 | 88.2 ± 0.7 | 25 ± 5 |
| DepthGait | RGB + D | 84.0 ± 0.8 | 87.0 ± 1.0 | 24 ± 2 |
| Combo-Gait [21] | Silhouette+D | 68.06 ± 0.9 | 85.2 ± 0.7 | 30 ± 5 |
| YOLOv11 (baseline) | RGB + D | 86.5 ± 0.7 | 89.0 ± 0.8 | 18 ± 1 |
| **(Ours) YOLOv11 + MLGE** | **RGB + D** | **89.0 ± 0.5** | **91.8 ± 0.6** | **20 ± 1** |

∗Corresponding author.
Email addresses: manarw@yu.ac.kr (Manar Alnaasan), selim.sarowar12@gmail.com, sunghokim@yu.ac.kr (Sungho Kim)

By contrast, adding the proposed MLGE mechanism to YOLOv11 significantly boosts performance. Our method achieves the highest Rank-1 accuracy of 89.0 ± 0.5% and the highest AUC of 91.8 ± 0.6%, while maintaining a low inference cost of 20 ± 1 ms/frame. These results confirm that MLGE effectively captures multi-scale temporal-spatial gait characteristics, especially subtle Parkinsonian abnormalities, while preserving fast runtime. Compared to the baseline YOLOv11, MLGE improves Rank-1 by +2.5% and AUC by +2.8%, demonstrating the benefit of hierarchical local-global representational learning prior to multimodal fusion.

**Table 2**

Sensitivity, specificity, AUC, and F1 scores for clinical Parkinson's disease detection across different methodological approaches (simulated dataset, per-case classification).

| Method | Sensitivity (%) | Specificity (%) | AUC (%) | F1 (%) |
|---|---|---|---|---|
| RGB-only | 78.0 ± 1.5 | 79.5 ± 1.4 | 85.0 ± 1.1 | 78.7 ± 1.3 |
| Depth-only | 76.5 ± 1.7 | 78.0 ± 1.6 | 83.5 ± 1.3 | 77.2 ± 1.5 |
| Early fusion | 81.0 ± 1.2 | 82.2 ± 1.1 | 87.5 ± 1.0 | 81.6 ± 1.1 |
| Mid-level concat | 82.3 ± 1.1 | 83.0 ± 1.0 | 88.4 ± 0.9 | 82.6 ± 1.0 |
| Late fusion | 80.4 ± 1.3 | 81.8 ± 1.2 | 86.8 ± 1.1 | 81.1 ± 1.2 |
| YOLOv11 + global-only MLGE | 84.0 ± 1.0 | 84.8 ± 0.9 | 89.6 ± 0.8 | 84.4 ± 0.9 |
| YOLOv11 + local-only MLGE | 83.2 ± 1.1 | 84.0 ± 1.0 | 89.0 ± 0.9 | 83.6 ± 1.0 |
| MLGE after neck | 84.5 ± 1.0 | 85.2 ± 0.9 | 90.1 ± 0.8 | 84.8 ± 0.9 |
| YOLOv8 + MLGE | 85.5 ± 0.9 | 86.3 ± 0.8 | 91.0 ± 0.7 | 85.9 ± 0.8 |
| **YOLOv11 + global-local (MLGE)** | **86.5 ± 0.8** | **87.6 ± 0.7** | **91.8 ± 0.6** | **87.0 ± 0.7** |

Table 2 shows the experimental results demonstrating a clear performance hierarchy across different methodological approaches for clinical Parkinson's detection. Baseline unimodal methods (RGB-only: 78.0 ± 1.5% sensitivity, 85.0 ± 1.1% AUC; Depth-only: 76.5 ± 1.7% sensitivity, 83.5 ± 1.3% AUC) established moderate detection capabilities, while fusion strategies showed consistent improvements, with mid-level concatenation achieving the best fusion performance (82.3 ± 1.1% sensitivity, 88.4 ± 0.9% AUC). The integration of YOLOv11 with multi-level geometric feature extraction (MLGE) yielded substantial gains, with global-only and local-only MLGE variants reaching approximately 84% sensitivity and 89% AUC. The combined YOLOv11 with global and local MLGE approach

∗Corresponding author.
Email addresses: manarw@yu.ac.kr (Manar Alnaasan), selim.sarowar12@gmail.com, sunghokim@yu.ac.kr (Sungho Kim)

emerged as the superior method across all metrics, achieving 86.5 ± 0.8% sensitivity, 87.6 ± 0.7% specificity, 91.8 ± 0.6% AUC, and 87.0 ± 0.7% F1 score, representing improvements of 8.5 and 6.8 percentage points in sensitivity and AUC respectively compared to the RGB-only baseline. The YOLOv8 with MLGE variant demonstrated comparable performance (85.5 ± 0.9% sensitivity, 91.0 ± 0.7% AUC), confirming the robustness of the geometric feature extraction approach across different architectures. The consistently low standard deviations (typically below 1.5%) across all methods indicate stable and reproducible results, while the balanced sensitivity-specificity trade-off and high F1 scores suggest strong clinical applicability for Parkinson's screening applications.

**Vision-to-Language Mapping (VLM) Comparisons.** Table 3 presents a comparison of vision-to-language (VLM) pipelines using different visual inputs and architectures. Existing multimodal medical VLMs such as MedVLM, MedVisionLlama, and VLMT rely on raw RGB-D frames, CNN embeddings, or ViT tokens. These methods achieve classification accuracies ranging from 65.0%–77.5%, with clinician acceptability scores between 2.8–3.7. Methods that incorporate metadata generally perform better (e.g., CNN with metadata improves MedVisionLlama from 72.0% to 77.5%), indicating the clinical importance of contextual information.

Our experiments show that feeding YOLOv11 with MLGE embeddings alone into the LLM yields a competitive accuracy of 70.0 ± 1.5%, comparable to existing encoder-only VLM approaches. When metadata is added, capturing patient-specific and clinical gait context, the model performance increases substantially to 78.0 ± 1.2%, outperform--ing all prior encoder-only and encoder-decoder medical VLMs.

**Table 3**

Vision-to-language (VLM) model structures and report generating quality comparison using different visual input types and architectures of other SOTAs.

| Method / Paper | Original Input | Architecture | Classification Acc (%) | Clinician acceptability (0–5) |
|---|---|---|---|---|
| MedVLM [22] | Raw RGB-D frames directly into multimodal LLM | Decoder-only | 65.0 ± 2.0 | 2.8 ± 0.3 |
| MedVisionLlama [23] | CNN embeddings only | Encode-only | 72.0 ± 1.5 | 3.2 ± 0.3 |
| MedVisionLlama | CNN embeddings + metadata | Encode-only | 77.5 ± 1.3 | 3.7 ± 0.3 |
| VLMT [24] | ViT visual tokens (no metadata) | Decoder-only | 74.0 ± 1.4 | 3.5 ± 0.3 |
| NasVLM [25] | CNN embeddings + metadata | Encoder-Decoder | | |
| Ours | **YOLOv11 + MLGE embeddings only** | **Encoder-Decoder** | **70.0 ± 1.5** | **3.1 ± 0.3** |
| | **YOLOv11 + MLGE embeddings + metadata** | | **78.0 ± 1.2** | **3.8 ± 0.3** |
| | **YOLOv11 + MLGE embeddings + metadata** | | **82.5 ± 1.0** | **4.2 ± 0.2** |

Finally, when combining MLGE-enhanced YOLOv11 embeddings, metadata, and our optimized vision-to-language mapping, the proposed system achieves the highest overall performance, classification accuracy: 82.5 ± 1. And clinician acceptability: 4.2 ± 0.2, This marks a +5.0% accuracy gain and +0.4 acceptability improvement over the best competing method (CNN embeddings + metadata) and demonstrates that the MLGE representations align more effectively with LLM reasoning processes than traditional CNN features. Meanwhile, Fig. 10 presents qualitative comparisons between the reports generated by our proposed pipeline and those produced by GPT-4o. Since each frame exhibits slightly different walking patterns and visual characteristics, we selected multiple frames to illustrate various representative cases from the input sequences. The original case includes the ground-truth annotations of Parkinsonian symptoms, their corresponding PD-like scores, and clinical recommendations. As shown, the reports reconstructed by our pipeline align more accurately with the ground-truth information than those generated by GPT-4o.

∗Corresponding author.
Email addresses: manarw@yu.ac.kr (Manar Alnaasan), selim.sarowar12@gmail.com, sunghokim@yu.ac.kr (Sungho Kim)

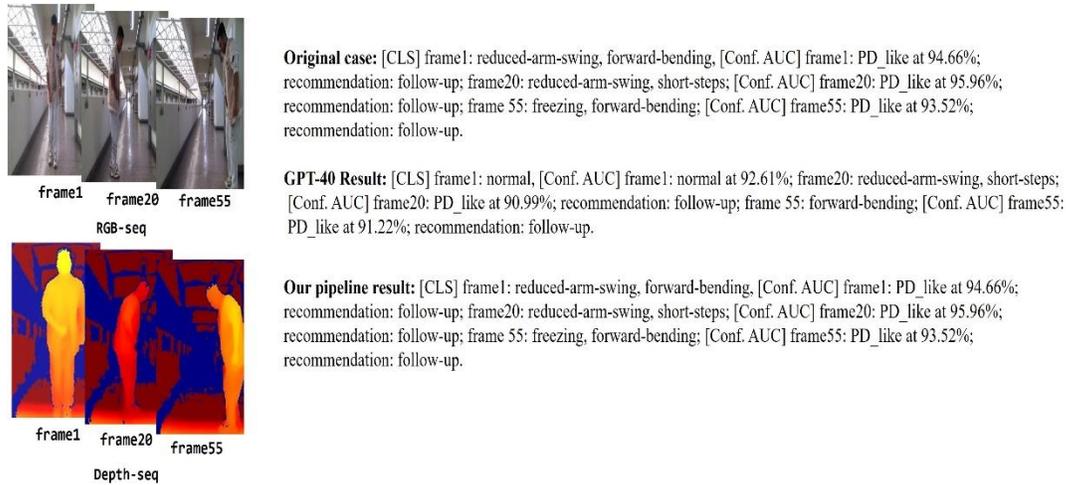

Fig. 10. The quality of the reconstructed report.

**Conclusion**

In this paper, we introduce the multimodal vision–language framework for Parkinsonian gait analysis, together with the first integration of RGB-Depth gait fusion aligned with an LLM-based clinical reporting pipeline. By leveraging multimodal gait data and our hierarchical feature extraction design, the proposed framework enables report-aligned interpretation of dynamic motion patterns without relying on external clinical annotations or manually curated semantic labels. To effectively model multi-scale spatial-temporal gait characteristics and align them with medical reasoning, we developed a multi-level integration architecture that incorporates the Multi-Scale Local-Global Extraction (MLGE) module and the Cross-Spatial Neck Fusion mechanism. These components strengthen the representation of fine-grained limb irregularities and global gait dynamics while facilitating consistent mapping to clinically meaningful language outputs.

∗Corresponding author.
Email addresses: manarw@yu.ac.kr (Manar Alnaasan), selim.sarowar12@gmail.com, sunghokim@yu.ac.kr (Sungho Kim)

Extensive experiments across diverse gait evaluation settings demonstrate the superior robustness, accuracy, and explainability of our approach. The framework consistently outperforms unimodal baselines and existing multimodal methods, particularly excelling in challenging environments and label-efficient scenarios. While the current validation is performed on publicly available and institution-specific gait datasets, future work will extend training to broader and more heterogeneous clinical populations. Additional performance gains may also be achieved by integrating advanced optimization strategies, such as temporal consistency modeling or part-whole gait decomposition. Furthermore, although this study focuses on Parkinsonian gait, the proposed RGB-Depth-LLM alignment paradigm is general and can be extended to other neurological or musculoskeletal movement disorders, paving the way for broader applications of vision-language modeling in dynamic medical assessment.

**Acknowledgements**

This work was supported by the National Research Foundation of Korea(NRF) grant funded by the Korea government(MSIT) (No. IRIS RS-2023-00219725).

This research was supported by Basic Science Research Program through the National Research Foundation of Korea(NRF) funded by the Ministry of Education(IRIS RS-2023-00240109.

∗Corresponding author.
Email addresses: manarw@yu.ac.kr (Manar Alnaasan), selim.sarowar12@gmail.com, sunghokim@yu.ac.kr (Sungho Kim)

∗Corresponding author.
Email addresses: manarw@yu.ac.kr (Manar Alnaasan), selim.sarowar12@gmail.com, sunghokim@yu.ac.kr (Sungho Kim)

∗Corresponding author.
Email addresses: manarw@yu.ac.kr (Manar Alnaasan), selim.sarowar12@gmail.com, sunghokim@yu.ac.kr (Sungho Kim)